\documentclass[letterpaper, 10 pt, conference]{ieeeconf}  % Comment this line out if you need a4paper

\IEEEoverridecommandlockouts                              % This command is only needed if 
\usepackage{graphics}
\usepackage{amsfonts}
\usepackage{amsmath}
\usepackage{bm}
\usepackage{upgreek}
\DeclareMathOperator*{\argmin}{\arg\!\min}
\usepackage{amssymb}
\usepackage{algorithm}
\usepackage{algpseudocode}
\usepackage{epsfig}
\usepackage{graphicx}
\usepackage{caption}
\usepackage{mathrsfs}
\usepackage{array}
\usepackage{multirow}
\usepackage{multicol}
\usepackage{booktabs}
\usepackage{makecell}
\usepackage{xcolor}
\usepackage{amsfonts}
\usepackage{flushend}
\usepackage{soul}
\usepackage[colorlinks, linkcolor=black]{hyperref}
\usepackage{stfloats}
\usepackage{balance}
\usepackage{threeparttable}

\newcommand\norm[1]{\lVert #1 \rVert}

\title{\LARGE \bf
CNS: Correspondence Encoded Neural Image Servo Policy
}

\author{Anzhe Chen$^{\#}$, Hongxiang Yu$^{\#}$, Yue Wang$^{*}$, Rong Xiong% <-this % stops a space
}

\begin{document}

\maketitle
\thispagestyle{empty}
\pagestyle{empty}

%%%%%%%%%%%%%%%%%%%%%%%%%%%%%%%%%%%%%%%%%%%%%%%%%%%%%%%%%%%%%%%%%%%%%%%%%%%%%%%%
\begin{abstract}

Image servo is an indispensable technique in robotic applications that helps to achieve high precision positioning. The intermediate representation of image servo policy is important to sensor input abstraction and policy output guidance. Classical approaches achieve high precision but require clean keypoint correspondence, and suffer from limited convergence basin or weak feature error robustness. Recent learning-based methods achieve moderate precision and large convergence basin on specific scenes but face issues when generalizing to novel environments. In this paper, we encode keypoints and correspondence into a graph and use graph neural network as architecture of controller. This design utilizes both advantages: generalizable intermediate representation from keypoint correspondence and strong modeling ability from neural network. Other techniques including realistic data generation, feature clustering and distance decoupling are proposed to further improve efficiency, precision and generalization. Experiments in simulation and real-world verify the effectiveness of our method in speed (maximum 40fps along with observer), precision (\textless 0.3° and sub-millimeter accuracy) and generalization (sim-to-real without fine-tuning). 
%Our code is available at \url{https://github.com/hhcaz/CNS}.
% Project homepage (full paper with supplementary text, video and code): \url{https://github.com/hhcaz/CNS}.
Project homepage (full paper with supplementary text, video and code): \url{https://hhcaz.github.io/CNS-home}.
\end{abstract}

%%%%%%%%%%%%%%%%%%%%%%%%%%%%%%%%%%%%%%%%%%%%%%%%%%%%%%%%%%%%%%%%%%%%%%%%%%%%%%%%

\section{Introduction}
\label{sec:intro}

%Image servo is an important branch of visual servoing. Given a desired image, the robot needs to adjust its pose to make the current image consistent with the desired image (Fig.\ref{fig: teaser}). It helps robots achieve high precision positioning, which is an indispensable technique in applications such as robotic manipulation \cite{levine2016end,zhong2019practical,puang2020kovis,paradis2021intermittent}. 

Image servo is an indispensable technique for robotic applications such as navigation and manipulation \cite{levine2016end,zhong2019practical,puang2020kovis,paradis2021intermittent}. Given the current image, the robot needs to adjust its pose to make it consistent with the desired image to achieve high precision positioning.

Traditional methods extract keypoint correspondence between current and desired images, and derive velocity control directly from  keypoints error (image-based visual servoing, IBVS) or estimate the relative pose transformation with extra object model (position-based visual servoing, PBVS). These methods achieve high servo precision, but suffer from either small convergence basin and error correspondence (IBVS), or imprecise object model and camera intrinsic (PBVS) \cite{kermorgant2011combine,Jin2022vga}. With the development of deep learning, it has become a trend to empower image servo with data.

Image servo is a sensory-act process, its intermediate representation is important in sensor input abstraction and policy output guidance. From this perspective, IBVS adopts keypoint correspondence as explicit representation. Following this line, prior works \cite{Jin2022vga,adrian2022dfbvs,harish2020dfvs,katara2021deepmpcvs} try to improve the accuracy and density of correspondence, or fine-tune the parameter of IBVS to improve the overall performance of IBVS based methods. However, these methods cannot overcome the intrinsic problem of IBVS.
Another line of methods \cite{saxena2017exploring,bateux2018training,yu2019siamese,felton2021siame} investigate the implicit representations. They encode image to latent vectors which naturally avoid explicit error correspondence and predict velocity in an end-to-end manner supervised by PBVS, which has a larger convergence basin. These methods converge well and achieve comparable precision with IBVS in specific scenes seen in training, however, cannot generalize well to novel scenes because some spurious scene-specific features are learned.

% \begin{figure}[t]
% \includegraphics[width=\linewidth]{figures/teaser0904.pdf}
% \centering
% \caption{We model keypoints and their correspondence as a graph and employ graph neural networks to predict velocity control for servoing.}
% \label{fig: teaser}
%   % \vspace{-0.5cm}
% \end{figure}

\begin{figure}[t]
\includegraphics[width=0.95\linewidth]{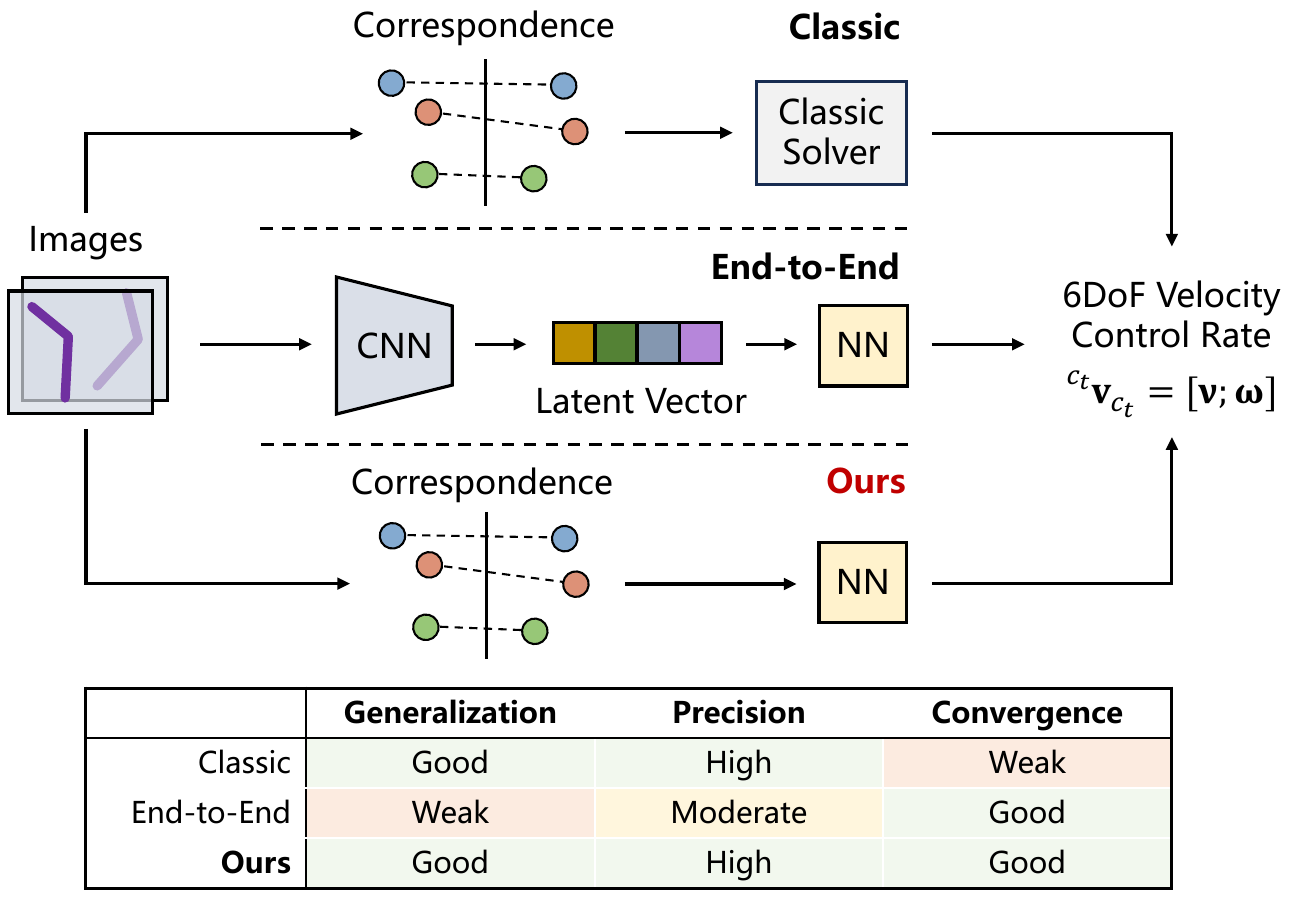}
\centering
\caption{We utilize explicit correspondence and neural policy, endowing image servo with generalization, high precision and large convergence basin.}
\label{fig: teaser}
  \vspace{-0.5cm}
\end{figure}

In this paper, we introduce Correspondence encoded Neural image Servo policy (CNS), which utilizes explicit correspondence with neural policy to combine the both advantages (Fig.\ref{fig: teaser}). We design the architecture, randomization and introduce several techniques to overcome the problems of explicit correspondence and neural policy to achieve:
\begin{itemize}
    \item High precision: We model arbitrary number of keypoints and the intermittent correspondence as a graph with time-variant structure and intuitively, CNS is built on a graph neural network (GNN). We use clustering and attentional aggregation to deal with error correspondence. We also simulate the error correspondence when randomization in training as data augmentation to further improve the error tolerance;
    \item Large convergence basin: Our neural policy is supervised by PBVS which intrinsically has larger convergence basin than IBVS. Moreover, we introduce graph convolutional gated recurrent unit to implicitly model scene structure which further improves the convergence and robustness to intermittent correspondence;
    \item Generalization: As keypoint correspondence isolates the appearance of the image from the neural policy, our model naturally generalizes to novel scenes. Besides, we predict a distance decoupled velocity which prevents the neural policy overfitting to scenes of specific scale that used in training. Several randomization techniques are also introduced for further improving the generalization.
\end{itemize}

We verify the effectiveness of CNS both in simulation and real-world. The policy trained in simulation can be directly transferred to novel scenes in real-world without any fine-tuning, relieving the burden of deployment.

\section{Related Works}

% \begin{figure*}[t]
% \includegraphics[width=\textwidth]{figures/what_is_vs.pdf}
% \centering
% \caption{The general procedure of image servo (left), current methods (middle), and the graph representation of keypoints and correspondence on this simple case using our method (right).}
% \label{fig: waht_is_vs}
%   \vspace{-0.5cm}
% \end{figure*}

\textbf{Classical Visual Servo:} Traditional visual servo includes IBVS, PBVS and hybrid approaches. IBVS \cite{chaumette2006visual,allibert2010predictive} use geometric features such as keypoint correspondence
extracted from images which is robust to calibration and model errors. However, it has limited convergence basin (feature loss problem \cite{Jin2022vga}, unpredictable 3D trajectory \cite{kermorgant2011combine}, Jacobian singularities or local minima \cite{chaumette2007potential}). PBVS uses camera's 3D poses as features and is globally asymptotically stable. However, it requires precise camera intrinsic and 3D models of observed objects for pose estimation. 
%Hybrid approaches combine the two methods. \cite{gans2007switch} switches between IBVS and PBVS to prevent the drawback of single controller. \cite{malis19992} replaces some of the 3D features by 2D information. \cite{kermorgant2011combine} sets PBVS as the core of the scheme and adds 2D information when necessary. Overall, these methods are general and usually highly precise with clean correspondence, but limited either in convergence basin or robustness to feature error.
Hybrid approaches switch between \cite{gans2007switch} or combine \cite{kermorgant2011combine,malis19992} the two methods to utilize the both advantages. Overall, these methods are general and usually highly precise with clean correspondence, but limited either in convergence basin or robustness to feature error.

\textbf{Learning Based Visual Servo:} To improve model's robustness and convergence, learning based methods are introduced to image servo. These methods can be classified into three categories. 

The first category tries to improve the observer, that is, the quality (correctness, density) of keypoint correspondence \cite{adrian2022dfbvs} with modern learning based feature matching methods \cite{detone2018superpoint,sarlin2020superglue,ni2023pats}, or optical-flow estimation methods \cite{harish2020dfvs,katara2021deepmpcvs}, and still use classic controller for servoing. These methods can generalize to novel scenes but do not overcome the intrinsic problem of controllers.

The second category focuses on improving the controller or the both. Given 3D mesh of objects, \cite{puang2020kovis} trains the neural observer on specific scenes to provide accurate keypoints for the neural controller to achieve high precision servoing. However, the trained observer and controller work only for seen scenes and fixed desired pose. \cite{yu2023hyper} treats the arbitrary desired pose servoing as a multi-task learning problem and use a hyper-net to generate weights for neural controller according to the given desired pose. However, it is still not general enough since the number of keypoints is fixed and it can not handle temporarily missing keypoints. 

The last category doesn't strictly distinguish between observers and controllers but predicts velocity or pose in an end-to-end manner. They \cite{saxena2017exploring,bateux2018training,yu2019siamese,felton2021siame} encode the current and desired images into latent vectors and predict the velocity or pose with a MLP. These methods could achieve comparable precision with classic methods and is robust to image occlusions or other feature error, however, cannot generalize to novel scenes since they adopt a pure convolution structure which is translation invariant intrinsically but not suitable for estimating rotation.

Our CNS takes matched keypoints provided by any detector-based feature matching methods (\emph{e.g.}, SIFT, ORB, AKAZE, SuperGlue \cite{sarlin2020superglue}) and uses GNNs to model arbitrary number of keypoints and intermittent correspondence. CNS benefits from existing generalizable observers with a compatible network architecture as well as error tolerant modeling ability from deep learning.

\textbf{Graph Neural Network:} Graph neural network is popular in point cloud processing \cite{qi2017pointnet,qi2017pointnet++,wang2019dynamic,landrieu2018large,wang2019graph,zhao2021point} for its permutation-invariant nature to directly handle irregular data. Yet various mechanisms have been developed to improve the performance of GNNs. We briefly introduce relevant structures used in our work. \cite{wang2019dynamic} proposes EdgeConv to perform graph convolution on kNN graphs. \cite{zhao2021point} incorporates the self-attention to attentionally aggregates neighbor node embeddings. \cite{cai2021graphnorm} introduces GraphNorm to accelerate training GNNs. %To the best of our knowledge, we are the first to employ GNN for visual servoing.
 
%===============================================================================

\section{Architecture}
\label{sec:methods}

\begin{figure}[t]
\includegraphics[width=\linewidth]{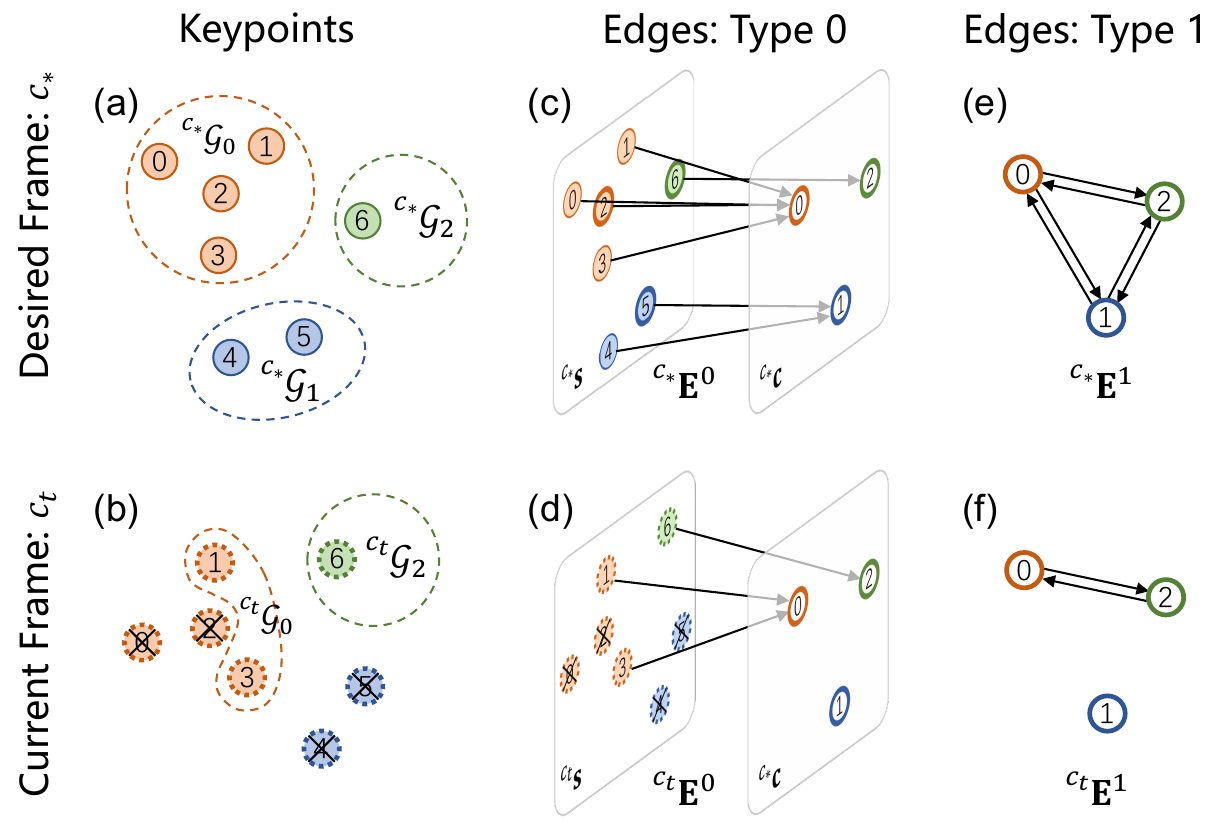}
\centering
\caption{A simple case illustration of the graph representation of 7 keypoints and correspondence. Keypoints are clustered into groups according to positions. Edges are categorized into 2 types for intra-cluster aggregation and inter-cluster information mutation.}
\label{fig: case_illustration}
  \vspace{-0.5cm}
\end{figure}

\subsection{Keypoints and Correspondence as Graph}
\label{sec: kp_graph}

Enrolling GNN for visual servo starts with the graph representation of keypoints and their correspondence. We assign grpah nodes' positions and initial embedding with the keypoint positions in normalized image plane and categorize edges into two types: one for intra-cluster embedding aggregation and the other for inter-cluster information mutation. 

\textbf{Notation:} To formally define the graph, we first define notations. The left superscript $c_{*}$ means variables obtained from desired pose, while $c_{t}$ from current pose. $\mathbf{s}$ represents keypoint positions in the normalized image plane. $^{c_*}\mathbf{s}_i$ represents the $i$-th keypoint of total $N$ keypoints in desired pose. $\mathcal{G}$ represents a group containing keypoint indices belonging to a specific cluster (we use Affinity-Propagation algorithm for clustering right after the extraction of $^{c_*}\mathbf{s}$). $^{c_*}\mathcal{G}_i$ contains indices of $i$-th cluster (total $N_c$ clusters) in desired pose. $\mathbf{c}$ represents the center keypoint of a cluster. Cluster center keypoint $\mathbf{c}_i$ for $i$-th cluster is chosen as the point closest to the mean keypoints position of that cluster: $\mathbf{c}_i = \argmin_{\mathbf{s}_j} \norm{\mathbf{s}_j - \frac{1}{N_{c_i}} \sum_{k\in \mathcal{G}_i} \mathbf{s}_k }$, where $N_{c_i}$ is the number of contained keypoints of $i$-th cluster. $\mathbf{E}$ represents edges where $\mathbf{E}[i][j] = 1$ means a directed connection from $j$-th source node to $i$-th target node. Node's position is chosen from $\mathbf{s}$ or $\mathbf{c}$.

\textbf{Case Illustration:} 
We present the graph structure by referring to a case shown in Fig.\ref{fig: case_illustration}. Suppose we have detected total 6 keypoints (circle with solid edge filled with a number indicating its index) in desired pose (Fig.\ref{fig: case_illustration}a). The keypoints are grouped into 3 clusters: $^{c_*}\mathcal{G}_0 = \{0, 1, 2, 3\}$ (red circles), $^{c_*}\mathcal{G}_1 = \{4, 5\}$ (blue circles), and $^{c_*}\mathcal{G}_2 = \{6\}$ (green circles). For $^{c_*}\mathcal{G}_0$, $^{c_*}\mathbf{s}_2$ is chosen as the center point $^{c_*}\mathbf{c}_0$. For $^{c_*}\mathcal{G}_1$, since $^{c_*}\mathbf{s}_4$ and $^{c_*}\mathbf{s}_5$ have equal distances to the mean position, we just random pick one as center point, and here we choose $^{c_*}\mathbf{s}_5$. For $^{c_*}\mathcal{G}_2$, it contains only one keypoint, therefore the center is chosen as that keypoint $^{c_*}\mathbf{s}_6$. Edges $^{c_*}\mathbf{E}^0$ define connections from each keypoint to center keypoint of its belonging cluster (Fig.\ref{fig: case_illustration}c), edges $^{c_*}\mathbf{E}^1$ define connections among each cluster center keypoints (Fig.\ref{fig: case_illustration}e):

\vspace{-0.3cm}
\begin{small}
\begin{equation}
    ^{c_*}\mathbf{E}^0 = \begin{bmatrix}
        1 & 1 & 1 & 1 & 0 & 0 & 0 \\
        0 & 0 & 0 & 0 & 1 & 1 & 0 \\
        0 & 0 & 0 & 0 & 0 & 0 & 1 \\
    \end{bmatrix}, \ 
    ^{c_*}\mathbf{E}^1 = \begin{bmatrix}
        0 & 1 & 1 \\
        1 & 0 & 1 \\
        1 & 1 & 0 \\
    \end{bmatrix}
\end{equation}
\end{small}
% For $^{c_*}\mathbf{E}^0$, source nodes are $^{c_*}\mathbf{s}$, target nodes are $^{c_*}\mathbf{c}$ (Fig.\ref{fig: case_illustration}c), while for $^{c_*}\mathbf{E}^1$, source nodes and target nodes are both $^{c_*}\mathbf{c}$ (Fig.\ref{fig: case_illustration}e).

\noindent where $^{c_*}\mathbf{E}^0$ connect $^{c_*}\mathbf{s}$ (source nodes) to $^{c_*}\mathbf{c}$ (target nodes) (Fig.\ref{fig: case_illustration}c), while for $^{c_*}\mathbf{E}^1$, source nodes and target nodes are both $^{c_*}\mathbf{c}$ (Fig.\ref{fig: case_illustration}e).

In current pose, suppose the corresponding keypoints of $^{c_*}\mathbf{s}_0$, $^{c_*}\mathbf{s}_2$, $^{c_*}\mathbf{s}_4$ and $^{c_*}\mathbf{s}_5$ are missing (Fig.\ref{fig: case_illustration}b). As a result, positions and node embeddings of $^{c_t}\mathbf{s}_0$, $^{c_t}\mathbf{s}_2$, $^{c_t}\mathbf{s}_4$ and $^{c_t}\mathbf{s}_5$ are meaningless, they shouldn't participate in the intra-cluster embedding aggregation. Therefore, we drop them from the index groups $\mathcal{G}$ and edges $\mathbf{E}^0$, yielding $^{c_t}\mathcal{G}_0 = \{1, 3\}$, $^{c_t}\mathcal{G}_1 = \varnothing$, $^{c_t}\mathcal{G}_2 = \{6\}$. Since none of keypoints in $\mathcal{G}_1$ is observed, the cluster embedding on center keypoint is meaningless and shouldn't participate in inter-cluster information mutation, we need to drop them from edges $\mathbf{E}^1$, yielding:

\vspace{-0.3cm}
\begin{small}
\begin{equation}
    ^{c_t}\mathbf{E}^0 = \begin{bmatrix}
        0 & 1 & 0 & 1 & 0 & 0 & 0 \\
        0 & 0 & 0 & 0 & 0 & 0 & 0 \\
        0 & 0 & 0 & 0 & 0 & 0 & 1 \\
    \end{bmatrix}, \
    ^{c_t}\mathbf{E}^1 = \begin{bmatrix}
        0 & 0 & 1 \\
        0 & 0 & 0 \\
        1 & 0 & 0 \\
    \end{bmatrix}
\end{equation}
\end{small}

\noindent Note that we still use $^{c_*}\mathbf{c}$ as target nodes for edges $^{c_t}\mathbf{E}^0$ (Fig. \ref{fig: case_illustration}d), because the cluster center keypoint at current pose may not be observed either. The source nodes and target nodes are also $^{c_*}\mathbf{c}$ for edges $^{c_t}\mathbf{E}^1$ (Fig.\ref{fig: case_illustration}f). Fig.\ref{fig: traj_of_graph} shows the evolution of graph structure in real-world experiments.

\textbf{Clustering for Efficiency:} If no clustering is applied, each keypoint becomes an individual cluster, we then have $^{c_*}\mathbf{E}^0 = \mathbf{I}_{N\times N}$ and $^{c_*}\mathbf{E}^1 = 1 - \mathbf{I}_{N\times N}$. In such condition, any graph convolution takes $^{c_*}\mathbf{E}^0$ is simply a MLP without aggregation, and $^{c_*}\mathbf{E}^1$ defines dense connected edges which consumes much memory and time for graph convolution when numerous keypoints are detected. Thus clustering obviously saves memory and inference time.

\textbf{Clustering for Higher Precision:} If no clustering is applied, each keypoint contributes equally to the final control rate, so does the noisy and mismatched keypoint. While clustering are used to achieve intra-cluster embedding aggregation by an attentional graph convolution (later described in section \ref{sec: nn_structure}), which has the possibility to lower the contribution of these keypoints.

\subsection{Network Architecture}
\label{sec: nn_structure}

\begin{figure}[tb]
\includegraphics[width=\linewidth]{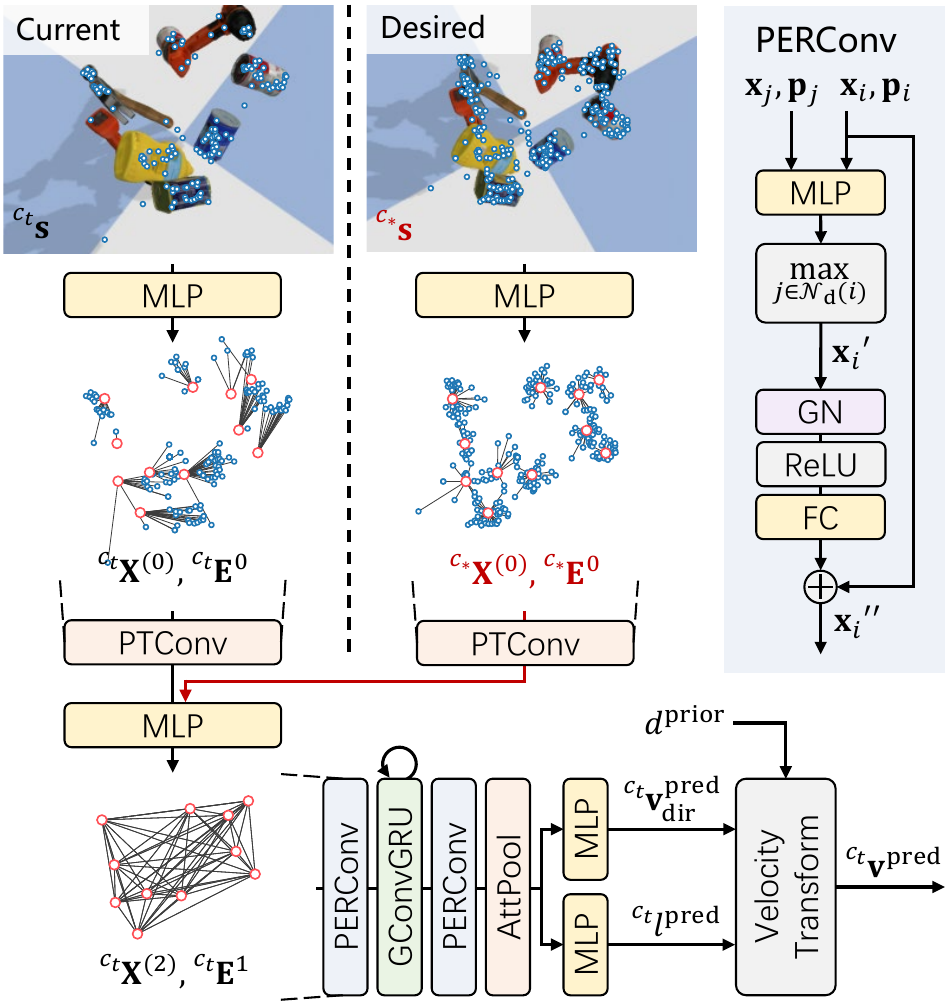}
\centering
\caption{Architecture of CNS. CNS takes over spatio-temporal features of keypoints and predicts distance decoupled velocity. PTConv aggregates intra-cluster embeddings with edge $^{c_t}\mathbf{E}^0$ or $^{c_*}\mathbf{E}^0$, PERConvs are used to achieve inter-cluster information mutation with edge $^{c_t}\mathbf{E}^1$. GConvGRU utilizes historical observations to implicitly model scene structure, brining larger convergence basin than IBVS and generating more smooth control.}
\label{fig: cns}
  \vspace{-0.5cm}
\end{figure}

By encoding keypoints and correspondence as a graph, we naturally build the policy with GNNs (Fig.\ref{fig: cns}).
To achieve intra-cluster node embedding aggregation, we use PointTransformer convolution \cite{zhao2021point} (denote as PTConv in Fig.\ref{fig: cns}) to aggregate features of keypoints at desired pose and current pose. The aggregated features are subtracted, concatenated and fused by a MLP.
Since low grain features have been aggregated to cluster centers, further convolutions can be conducted only on these centers which significantly reduce the computing complexity when numerous keypoints are encountered. We propose to use two point-edge-residual convolution (PERConv) layers to propagate information among cluster centers. %As ablation study results on clustering (in Supplementary Material, section \ref{sec: ablation_net_structure}) shows, clustering boosts the efficiency and precision.

\textbf{Graph Convolutional Gated Recurrent Unit:}
Keypoints provided by observers may be temporarily missing and jitter in image plane. A pure feedforward network structure may suffer from noises in a sequential decision task, therefore, it's natural to incorporate temporal modeling ability in our controller. We thus propose graph convolutional GRU (GConvGRU) built upon PERConv and GRU as Eq.\ref{GConvGRU} (Supplementary Material, section \ref{subsec: strcuture_of_gconvgru}). Our GConvGRU is different from gated graph convolution proposed by \cite{li2015gated}. The latter runs graph convolution and GRU cell sequentially, while in our GConvGRU, graph convolution directly participates in the predictions of gates and the hidden state. %As ablation study on GRU structure (in Supplementary Material, section \ref{sec: ablation_net_structure}) shows, our structure achieves higher success ratio and precision than \cite{li2015gated} and CNS without GRU.

\textbf{Network Predicition:}
CNS predicts distance decoupled velocity for better generalization. Specifically, it predicts the direction $\mathbf{v}^{\rm pred}_{\rm dir} = [{\bm \upnu}^{\rm pred}_{\rm dir}; {\bm \upomega}^{\rm pred}_{\rm dir}]$ and norm $\mathcal{T}(l^{\rm pred})$ of the distance decoupled velocity. Following the IBVS requiring depth value (can be roughly estimated) to calculate velocity, we also need a distance prior (only a scalar) to scale the linear velocity of network predictions for actual control:

\vspace{-0.3cm}
\begin{small}
\begin{equation}
\label{raw_pred_tform}
    {\bm \upnu}^{\rm pred} = \frac{{\bm \upnu}^{\rm pred}_{\rm dir}}{\norm{\mathbf{v}^{\rm pred}_{\rm dir}}_2} \mathcal{T}(l^{\rm pred}) \cdot d ;~~
    {\bm \upomega}^{\rm pred} = \frac{{\bm \upomega}^{\rm pred}_{\rm dir}}{\norm{\mathbf{v}^{\rm pred}_{\rm dir}}_2} \mathcal{T}(l^{\rm pred})
\end{equation}
\end{small}

\noindent where $\mathcal{T}(\cdot) = 1 + {\rm ELU}(\cdot)$. $\mathcal{T}$ decays exponentially to zero when input is negative, encouraging the network to predict more accurate velocity when close to the desired pose to avoid damping behavior. When input is positive, $\mathcal{T}$ behaves as a linear function, preventing the network predicting exaggerated large velocity.

\subsection{Loss}
\label{subsec:objectives_design}
We use PBVS \cite{chaumette2006visual} as supervision since poses are always known in simulation. We divide the linear velocity of supervision by ground truth distance prior $d^{\rm gt} = \norm{ \ ^{c_*} \mathbf{p}_o}_2$ (distance from scene center to camera at desired pose) to obtain the distance decoupled representation of the supervision: $\mathbf{v}^{\rm gt}_{\rm dd} = [\bm{\upnu}^{\rm gt} / d^{\rm gt}; \bm{\upomega}^{\rm gt}]$. We supervise the direction and the norm of the distance decoupled velocity separately:
\begin{equation}
\label{servo_loss}
\begin{aligned}
    \mathcal{L}_{\rm dir} &= 1 - {\rm CosineSimilarity}(\mathbf{v}^{\rm pred}_{\rm dir}, \mathbf{v}^{\rm gt}_{\rm dd}) \\
    \mathcal{L}_{\rm norm} &= {\rm MSE} \left( l^{\rm pred}, \mathcal{T}^{-1} \left( \norm{\mathbf{v}^{\rm gt}_{\rm dd}}_2 \right) \right)
\end{aligned}
\end{equation}
% with the final servo loss as a weighted sum of them: $\mathcal{L}_{\rm servo} = \mathcal{L}_{\rm dir} + 0.1 \mathcal{L}_{\rm norm}$.
with the final servo loss as: $\mathcal{L}_{\rm servo} = \mathcal{L}_{\rm dir} + 0.1 \mathcal{L}_{\rm norm}$.

\section{Randomization in Training}
\label{sec: data_generation}
Thanks to the graph based representation, we need to only sample keypoints rather than render images for faster training. It also enables us to directly simulate the non-idealities of keypoints and correspondence regradless of detailed image appearance for generalization.
%More importantly, it enables the randomization of the correspondence, which improves the performance of the neural policy, as well as eliminates the need of sim-to-real transfer.

\subsection{Keypoints Distribution Simulation} 
\label{subsec: kp_distri}
\begin{figure}[tb]
\includegraphics[width=0.45\textwidth]{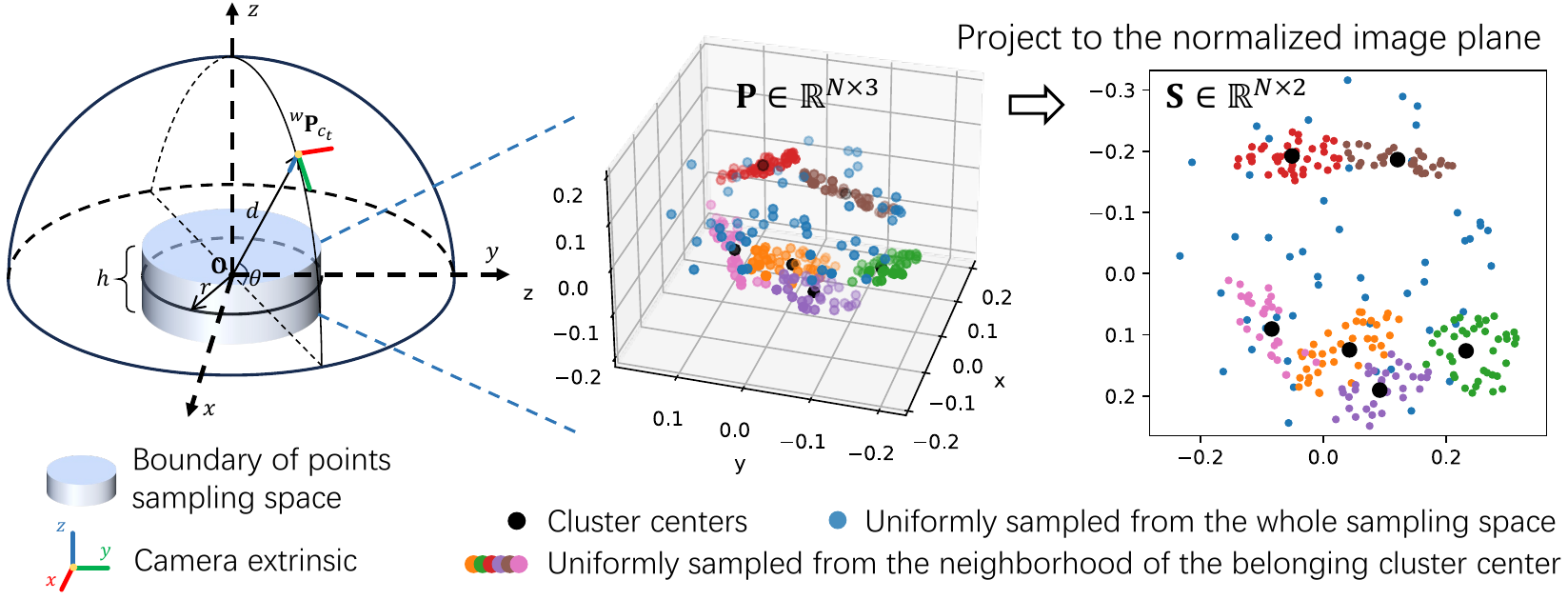}
\centering
\caption{
Keypoints sampling and projection. The blue cylinder defines the boundary of points sampling space in training, or ranges of positions of objects from YCB datasets \cite{calli2015ycb} in evaluation (in environment ${\rm E}_{\rm render}$ described later in section \ref{sec: sim_setup}). After sampling points in 3D space, we project them to the normalized image plane as the detected keypoints.
}
\label{fig: data_gen}
    \vspace{-0.5cm}
\end{figure}

Empirically, we assume the distribution of keypoints inversely projected to 3D space as a union of multiple bounded uniform distributions. Denoting $\mathcal{S}(n, x, y, z, h, a, b)$ as a randomization scheme to uniformly sample $n$ points in a elliptic cylinder located at center $(x, y, z)$ with height $h$, major axis $a$ and minor axis $b$. Given total number of points $N$ in scene with maximum scene boundary size $r$, we first determine the number of clusters to generate as $N_c$. Each cluster $i$ is randomly assigned with $n_i$ points with cluster center coordinates as $\{(x_i^c, y_i^c, z_i^c)\}_{i=1}^{N_c}$. Afterwards, points in $i$-th cluster are generated: $\mathbf{P}_i = \{(x_j, y_j, z_j)\}_{j=1}^{n_i} = \mathcal{S}(n_i, x_i^c, y_i^c, z_i^c, 0.1r, a, b)$. Random rotation around cluster's center is applied to all points belonging to the same cluster to mimic the various object surfaces in the scene. Finally, the residual points are uniformly distributed in the whole scene: $\mathbf{P}_0 = \mathcal{S}(N -\sum_i n_i, 0, 0, 0, 0.5r, r, r)$. The overall 3D points set is the union: 
%$\mathbf{P} \in \mathbb{R}^{N \times 3} = \mathbf{P}_0 \cup \left(\bigcup_{i=1}^{N_c} \mathbf{P}_i\right)$. 
$\mathbf{P} \in \mathbb{R}^{N \times 3} = \cup_{i=0}^{N_c} \mathbf{P}_i$.
Given camera's extrinsic $^c_w \mathbf{T}$ and intrinsic $\mathbf{K}$, keypoints in normalized image plane $\mathbf{S} \in \mathbb{R}^{N \times 2}$ can be obtained by projection: $\mathbf{S} = {\rm Project}(\mathbf{K}, \ ^c_w \mathbf{T}, \ \mathbf{P})$. Fig.\ref{fig: data_gen} shows an example of generated points and their projection in the normalized image plane. Detailed values of hyper-parameters to generate points are listed in Supplementary Material, section \ref{sec: values_of_hyper-params}.

\subsection{Non-ideal Correspondence} Only a subset of keypoints from desired pose frame can find their correspondence in current pose when facing object occlusions and large perspective change. Since the perspective change is resulted from camera motion, the initial observable region may shift to other areas when moving from initial pose to desired pose. To simulate the shifting observable region and missing keypoints behavior, we assign an observable probability $p_i$ and a jitter time constant $\tau_i$ to each keypoint $\mathbf{s}_i \in \mathbf{S}$. We define three Gaussian kernels in normalized image plane where keypoints close to kernel centers have higher $p_i$:
\begin{equation}
\label{obs_prob}
    p_i = \max(\{\kappa(\norm{\mathbf{s}_i - \mathbf{k}_j}_2, 0.3r)\}_{j=1}^{3})
\end{equation}
where $\kappa(\mu, \sigma) = \exp(-\mu^2 / (2\sigma^2))$, $\mathbf{s}_i=(x_i^{\rm kp}, y_i^{\rm kp})$ is the coordinate of $i$-th keypoint and $\mathbf{k}_j = (x_j^{\rm kc}, y_j^{\rm kc})$ is the coordinate of $j$-th kernel center. The kernel center changes with camera motion to produce observable region shifting behavior. This can be implemented by sampling $x_j^{\rm kc}$ and $y_j^{\rm kc}$ from 1D Perlin noise to realize smooth and random shifting trajectory:
\begin{equation}\label{region_shift}
    {x}_j^{\rm kc} = {\rm Perlin1D}\left(\frac{0.2}{\pi} \norm{(\theta \mathbf{u})|_{t_0}^{t_{\rm now}}}_2 + \frac{0.2}{r} \norm{\mathbf{t} |_{t_0}^{t_{\rm now}}}_2 \right)
\end{equation}
where $(\theta \mathbf{u})|_{t_0}^{t_{\rm now}}$ denotes axis-angle representation of relative rotation from initial simulation time step $t_0$ to current $t_{\rm now}$, $\mathbf{t} |_{t_0}^{t_{\rm now}}$ denotes the relative translation from $t_0$ to $t_{\rm now}$.

Given observable probability $p_i$ derived from Eq.\ref{obs_prob} and $\tau_i \sim \mathcal{U}(0.5, 5)$, a keypoint reverses its state via a kinetic Monte Carlo (KMC) process with average time from observable to missing $\tau_i^{\rm o2m} = p_i \tau_i$ and average time from missing to observable $\tau_i^{\rm m2o} = (1 - p_i) \tau_i$.

%===============================================================================

%===============================================================================

\begin{figure*}[t]
    \centering
    \includegraphics[width=\textwidth]{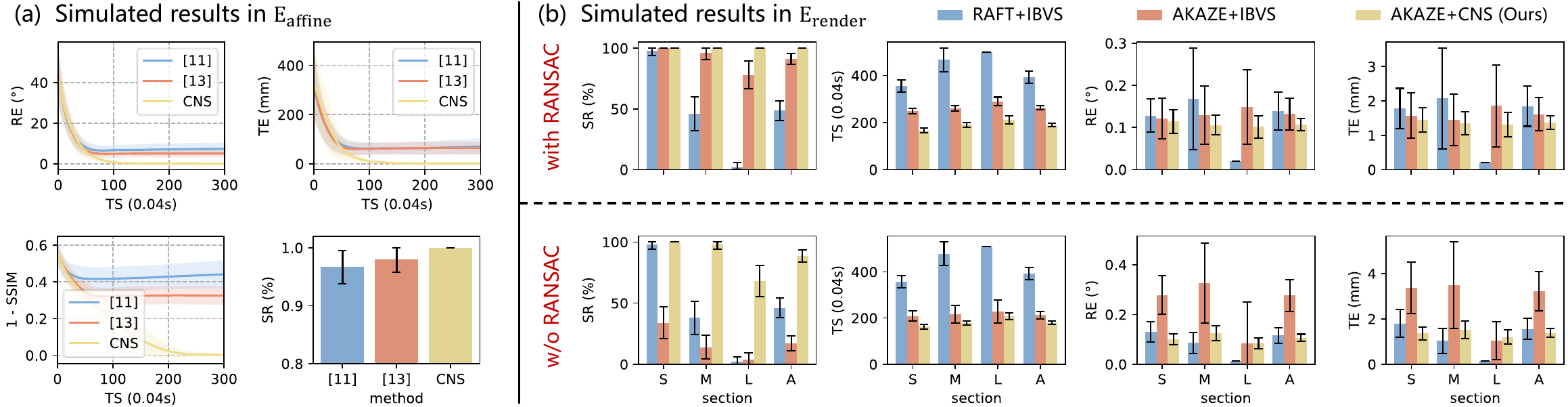}
    \caption{Comparisons in simulation. (a) Comparison with \cite{bateux2018training} and \cite{felton2021siame} in environment ${\rm E}_{\rm affine}$, with average initial RE of 44.69$^\circ$ and TE of 284.6mm. Each statistic is calculated from 150 runs. We assume a success servo episode when the final RE is less than 10° and TE less than 10cm. (b) Comparison with IBVS based methods in environment ${\rm E}_{\rm render}$ with or without RANSAC. Each statistic is calculated from 150 runs. "S", "M", "L" denote sections (each containing 50 scenes) of small, median and large initial RE (with average of 24.06$^\circ$, 67.38$^\circ$ and 136.46$^\circ$), respectively. "A" denotes all the scenes with average initial RE of 75.96$^\circ$ and TE of 247.9mm. We assume a success servo episode when the final RE is less than 3° and TE less than 3cm.}
    \label{fig: comparison_in_simulation}

\vspace{0.2cm}
\centering
\captionof{table}{Comparison with IBVS in real-world. Each statistic is calculated from 60 runs with average initial RE of 86.56$^\circ$ and TE of 148.8mm. Both IBVS and CNS use SIFT as the observer. We assume a success servo episode when the final RE is less than 3° and TE less than 3cm.}
\label{tab: comp_ibvs_real}

\vspace{0.2cm}
\resizebox{\linewidth}{!}{
\begin{tabular}{l|ll|ll|ll}
% \hline
\specialrule{.08em}{0pt}{0pt}
\multirow{2}{*}{}& \multicolumn{2}{l|}{Scene-Easy, with RANSAC}                   & \multicolumn{2}{l|}{Scene-Easy, w/o RANSAC}                             & \multicolumn{2}{l}{Scene-Hard, with RANSAC} \\ \cline{2-7}
                 & \multicolumn{1}{l|}{IBVS}               & CNS (ours)           & \multicolumn{1}{l|}{IBVS}               & CNS (ours)                    & \multicolumn{1}{l|}{IBVS}                 & CNS (ours) \\ 
% \hline
\specialrule{.08em}{0pt}{0pt}
SR (\%) (95\% ci) & \multicolumn{1}{l|}{98.33 (95.90, 100)} & \textbf{100}         & \multicolumn{1}{l|}{8.33 (1.34, 15.33)} & \textbf{75.00} (64.04. 85.96) & \multicolumn{1}{l|}{70.00 (58.40, 81.60)} & \textbf{96.67} (92.12, 100) \\ \hline
TS (0.04s)       & \multicolumn{1}{l|}{\textbf{145.3}±46.3}& 158.9±44.0           & \multicolumn{1}{l|}{358.8±65.1}         & \textbf{136.6}±49.4           & \multicolumn{1}{l|}{232.5±45.2}           & \textbf{217.8}±51.2     \\ \hline
RE (°)           & \multicolumn{1}{l|}{0.147±0.088}        & \textbf{0.053}±0.046 & \multicolumn{1}{l|}{1.368±0.429}        & \textbf{0.067}±0.033          & \multicolumn{1}{l|}{0.523±0.476}          & \textbf{0.101}±0.125     \\ \hline
TE (mm)          & \multicolumn{1}{l|}{0.843±0.505}        & \textbf{0.308}±0.162 & \multicolumn{1}{l|}{8.256±2.657}        & \textbf{0.385}±0.156          & \multicolumn{1}{l|}{2.839±2.867}          & \textbf{0.367}±0.438     \\ 
% \hline
\specialrule{.08em}{0pt}{0pt}
\end{tabular}
}
\vspace{-0.4cm}
\end{figure*}

\section{Experimental Results}
\label{sec:result}

\subsection{Simulation Environment Setup}
\label{sec: sim_setup}
We introduce two benchmark environments and stopping criterion for fairly benchmarking servo policies. The first environment named ${\rm E}_{\rm render}$ randomly places 8 to 12 objects from YCB datasets \cite{calli2015ycb} in a scene (Fig.\ref{fig: exp_setup}a in Supplementary Material). We generate total 150 scenes and randomly sample an initial pose and a desired pose for each scene. Image observed from specific pose is rendered by PyBullet simulation engine. The second environment named ${\rm E}_{\rm affine}$ use 2D image as scene (Fig.\ref{fig: exp_setup}b in Supplementary Material), similar as \cite{bateux2018training,felton2021siame} do. Since the scene is pure 2D, the resulting rendered image is affined from the original image. As for the stopping criterion, since keypoints are already extracted at current pose and desired pose for CNS and IBVS, we therefore use average keypoint position error as criterion and stop current servo process when error is below certain threshold and no longer decreases for 20 steps. While for \cite{bateux2018training} and \cite{felton2021siame}, we use structural similarity (SSIM) \cite{wang2004image} error between current image and desired image as criterion. 

\textbf{Metrics:} We use the following metrics to evaluate the performance of servo policies: (1) SR (success ratio, we also add 95\% confidence interval in tables), (2) TS (time steps to convergence, each time step lasts for 0.04s), (3) RE (rotation error), (4) TE (translation error), (5) mOT (mean time cost on observer in each frame), (6) mCT (mean time cost on controller in each frame), (7) mTT (mean total time cost on each frame, the summation of mOT and mCT).

\subsection{Simulation Results}
\label{sec: sim_results}

We benchmark our CNS and those implicit models in ${\rm E}_{\rm affine}$ with the same image used by \cite{bateux2018training} and \cite{felton2021siame}. All the models achieve high success ratio while our model achieves best servo precision (Fig.\ref{fig: comparison_in_simulation}a). The servo precision of \cite{bateux2018training} and \cite{felton2021siame} is limited because they use a MLP taking features from the last convolution layer of pre-trained CNN model (specifically trained for image classification) for servo. However, difference on the most coarse feature map struggles to capture minor difference from original image scale since the feature map has been down-sampled multiple times, resulting limited servo precision.

%We then replace the image with 150 images selected from MSCOCO2014 dataset to verify the generalization ability of \cite{bateux2018training} and \cite{felton2021siame}. Results show they are not able to generalize to unseen scenes, while our model still preserve high success ratio and high servo precision. Note in above experiments although we have obtained the keypoint positions, we still use SSIM criterion to stop servo for fair comparison. As \cite{bateux2018training} and \cite{felton2021siame} fail in unseen scenes, it's meaningless to discuss the time steps to convergence, we'd rather plot the average time-varing SSIM error and control rate predicted by models for an intuitive comparison.

We compare our CNS with IBVS in ${\rm E}_{\rm render}$. Scenes are divided into three sections (S, M, L) with increasing rotational deviation between initial and desired pose. Section A means the average result of all three sections. Results (Fig.\ref{fig: comparison_in_simulation}b) indicate that IBVS controller fails more on scenes with large initial RE. Compared with AKAZE+IBVS, RAFT\cite{teed2020raft}+IBVS performs worse when initial RE increases, we assume the reason is that data for training RAFT \cite{teed2020raft} may not contain such large rotation deviation, resulting weak generalization to these conditions. Our model success in all scenes and achieves highest servo precision in most scenes. Previous experiments use RANSAC as the additional post-processing step to reject outlier correspondences. When this geometric verification step is removed (Fig.\ref{fig: comparison_in_simulation}b, "w/o RANSAC"), AKAZE+IBVS shows significant performance drop while our model still preserves high success ratio and servo percision, which is more robust to mismatches. 

\begin{figure}[tb]

\begin{minipage}{\linewidth}
    \centering
    \includegraphics[width=\linewidth]{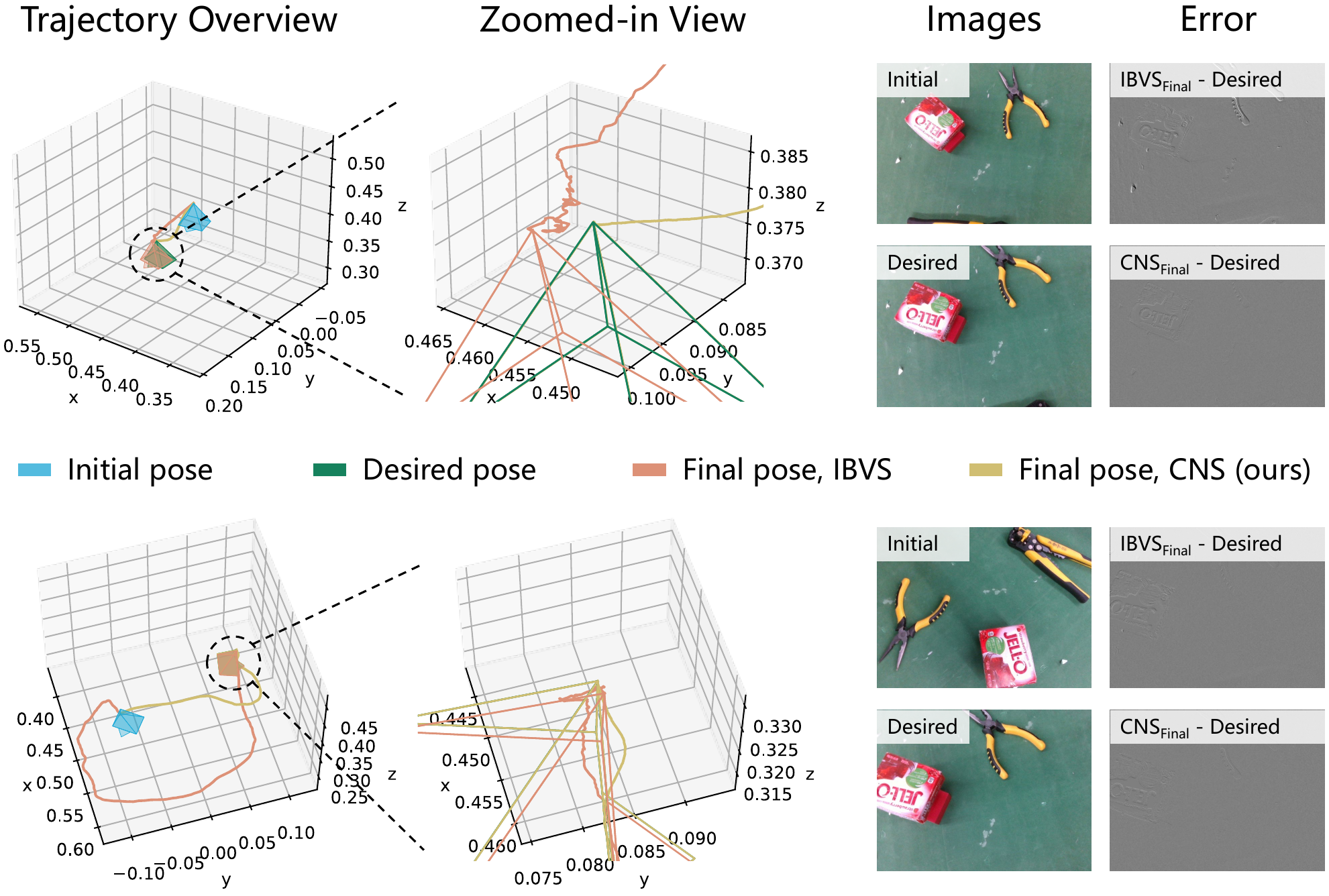}
    \caption{Sampled success cases of IBVS and CNS in real-world experiments with Scene-Hard setup. We visualize two cases. In each case, we plot the overview trajectories and their zoomed-in views around the desired pose. Next to the 3D trajectories are images obtained from initial and desired poses, and the gray-scale images representing the photometric error between the desired image and images obtained from the final poses guided by IBVS and CNS.}
\label{fig: exp_real_large}
\end{minipage}

  \vspace{0.3cm}

\begin{minipage}{\linewidth}
    \includegraphics[width=\linewidth]{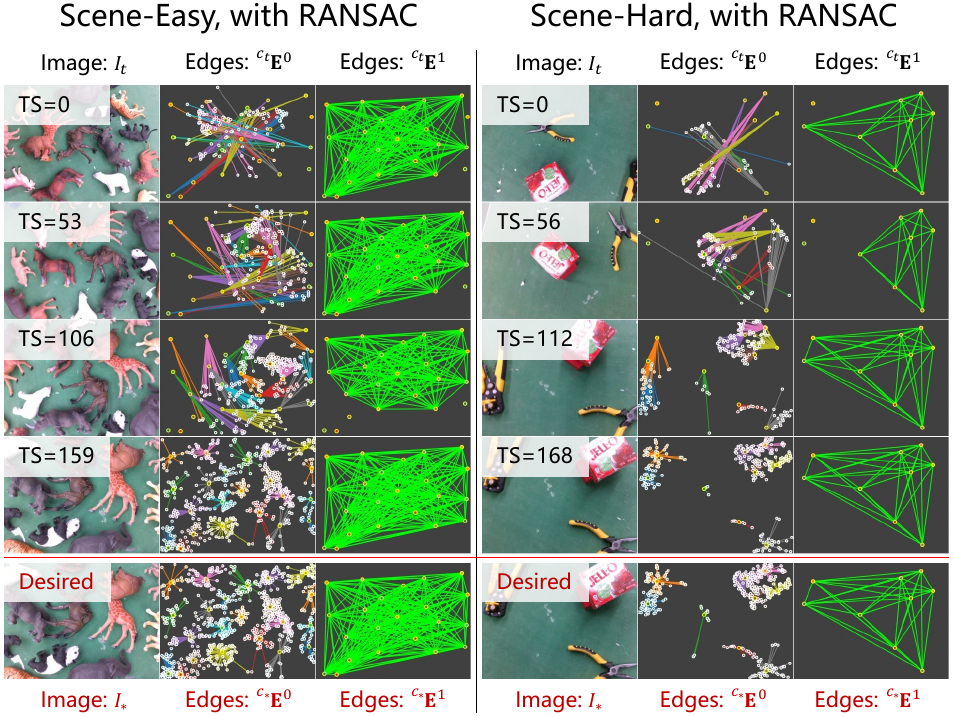}
    \centering
    \caption{Two samples of first perspective sequence of CNS in Scene-Easy (left) and Scene-Hard (right). In each scene, the first column shows images from current poses (except the last image from the desired pose), the second column shows the structure of edges $^{c_t}\mathbf{E}^0$ (except the last row to be $^{c_*}\mathbf{E}^0$) and the third column shows $^{c_t}\mathbf{E}^1$ (except the last row to be $^{c_*}\mathbf{E}^1$). In edges $^{c_t}\mathbf{E}^0$ or $^{c_*}\mathbf{E}^0$, points with white border are $^{c_t}\mathbf{s}$ and $^{c_*}\mathbf{s}$ respectively, while points with yellow border are always $^{c_*}\mathbf{c}$, as illustrated in section \ref{sec: kp_graph}.}
    \label{fig: traj_of_graph}
\end{minipage}

\vspace{-0.5cm}

\end{figure}

% \begin{table*}[t]
% \centering
% \caption{Ablation on network structure. Clustering boosts the convergence, precision and computing efficiency, and our GConvGRU further improves the convergence and precision.}
% \label{tab: ablation_net_structure2}
% % \resizebox{\linewidth}{!}{
% \begin{tabular}{l|l|l|l|l|l}
% \hline
%                                         & SR(\%) (95\% ci)     & TS (0.04s)  & RE (°)      & TE (mm)     & mCT (ms) \\ \hline
% CNS: Cluster + GConvGRU (ours)          & \textbf{100}          & 207.8± 43.8 & \textbf{0.115}±0.111 & \textbf{1.394}±1.402 & 11.51±1.65 \\ \hline
% CNS - Cluster                           & 99.33 (98.03, 100)   & 378.7±182.6 & 0.722±0.623 & 9.294±9.182 & 67.40±40.6 \\ \hline
% CNS - GConvGRU                          & 99.33 (98.03, 100)   & 218.6± 62.0 & 0.230±0.288 & 2.922±3.601 & \textbf{10.13}±1.40 \\ \hline
% CNS - GConvGRU + GGNN\cite{li2015gated} & 97.33 (94.76, 99.91) & \textbf{203.4}±118.3 & 0.241±0.283 & 2.847±3.905 & 10.17±1.38 \\ \hline
% \end{tabular}
% % }
% \end{table*}

\subsection{Real-world Environment Setup}
\label{sec: real_world_setup}
We compare our CNS with traditional high precision IBVS in two real-world scenes: Scene-Easy and Scene-Hard (Fig.\ref{fig: exp_setup}c in Supplementary Material). We uniformly scatter 20 objects (this results in rich keypoints) in Scene-Easy, and randomly sample 60 initial-desired pose pairs for statistics. In Scene-Hard, we use the same initial-desired pose pairs as those in Scene-Easy, but there are only 3 objects which are often partially observed at the initial or desired poses (Fig.\ref{fig: exp_real_large}). Moreover, the red box locates much higher than the other two yellow pliers.
% Results (Table.\ref{tab: comp_ibvs_real}) show CNS has a higher SR in Scene-Easy and achieves comparable precision with IBVS. 

\subsection{Real-world Results}
\label{sec: real_world_results}
Results show that CNS achieves higher success ratio and precision than IBVS in Scene-Easy (Table \ref{tab: comp_ibvs_real}).
In Scene-Hard, although CNS does not achieve 100$\%$ SR, it is much higher than IBVS. We also remove the RANSAC to inspect models' dependency on post-processing. Results show CNS is more robust to noisy and mismatched keypoints than IBVS while IBVS almost fails in all scenes. Moreover, CNS generate more smooth trajectories than IBVS (Fig.\ref{fig: exp_real_large}).

\begin{table}[tb]
\centering
\caption{Ablation study on network structure. Each statistic is obtained from 150 runs in simulation environment ${\rm E}_{\rm render}$. Clustering boosts the convergence time, precision and computing efficiency, and our GConvGRU further improves the convergence and precision. (-Cluster: replace PTConv with a point-wise MLP; -GConvGRU: replace GConvGRU with a point-wise MLP; +GGNN: replace GConvGRU with GGNN \cite{li2015gated}.)}
\label{tab: ablation_net_structure2}

% \begin{threeparttable}

    \resizebox{\linewidth}{!}{
    \begin{tabular}{l|l|l|l|l}
    % \hline
    \specialrule{.08em}{0pt}{0pt}
               & CNS (base) & -Cluster  & -GConvGRU  & +GGNN\cite{li2015gated}     \\ 
    % \hline
    \specialrule{.08em}{0pt}{0pt}
    SR (\%)    & \textbf{100} & 99.33     & 99.33      & 97.33     \\ \hline
    TS (0.04s) & 207.8±43.8 & 378.7±183 & 218.6±62.0 & \textbf{203.4}±118 \\ \hline
    RE (°)     & \textbf{0.12}±0.11  & 0.72±0.62 & 0.23±0.29  & 0.24±0.28 \\ \hline
    TE (mm)    & \textbf{1.39}±1.40  & 9.29±9.18 & 2.92±3.60  & 2.84±3.91 \\ \hline
    mCT (ms)   & 11.5±1.65  & 67.4±40.6 & \textbf{10.1}±1.40  & 10.2±1.38 \\ 
    % \hline
    \specialrule{.08em}{0pt}{0pt}
    \end{tabular}
    }
    
    % \begin{tablenotes}
    %     \footnotesize
    %     \item[*] -Cluster: replace PTConv with a point-wise MLP; 
    %     \item[*] -GConvGRU: replace GConvGRU with a point-wise MLP; 
    %     \item[*] +GGNN: replace GConvGRU with GGNN.
    % \end{tablenotes}

% \end{threeparttable}
\vspace{-0.4cm}
\end{table}

\subsection{Ablation Study on Network Structure}
\label{sec: ablation_net_structure2}
To evaluate the effectiveness of clustering, we set the structure illustrated in section \ref{sec: nn_structure} as base CNS model and replace the PTConv of base CNS with a point-wise MLP, resulting in a huge performance degradation in convergence time, servo precision and computing efficiency (Table \ref{tab: ablation_net_structure2}, column "-Cluster"). Compared with CNS without GConvGRU or using GGNN proposed by \cite{li2015gated}, the base model improves over 50\% in precision and reaches 100\% success rate in ${\rm E}_{\rm render}$, with a slight increase ($\sim$10\%) on network inference time (Table \ref{tab: ablation_net_structure2}, column "-GConvGRU" and "+GGNN").

\subsection{Discussion}
\label{sec: discussion}
Exhaustive experiments and ablation study are conducted to prove that CNS achieves:

\textbf{High precision:} CNS achieves \textless 0.3° and sub-millimeter precision in real-world experiments. It is much robust to error correspondence than IBVS (Table \ref{tab: comp_ibvs_real}, column "Scene-Easy, w/o RANSAC"). It also gains higher precision than end-to-end learning methods \cite{bateux2018training,felton2021siame} (Fig.\ref{fig: comparison_in_simulation}a). Ablation study on network architecture (section \ref{sec: ablation_net_structure2}) verifies the effectiveness of clustering on servo precision and convergence.

\textbf{Large convergence basin:} CNS successfully converges even with large initial pose deviation (Fig.\ref{fig: comparison_in_simulation}b, section "L"). It achieves higher success ratio than IBVS in real-world experiments (Table \ref{tab: comp_ibvs_real}) and end-to-end learning methods \cite{bateux2018training,felton2021siame} in simulation (Fig.\ref{fig: comparison_in_simulation}a). Ablation study on network architecture (section \ref{sec: ablation_net_structure2}) verifies the effectiveness of proposed GConvGRU on servo precision and convergence.

\textbf{Generalization:} Real-world experiments are conducted with model trained purely in simulation. The distance decoupled velocity design allows CNS workable to scenes of any scale (Supplementary Material, section \ref{sec: ablation_on_dd_vel}) at the cost of estimating an extra scalar. Ablation study on imprecise distance prior estimation (Supplementary Material, section \ref{subsec: distance_estimation}) shows the tolerance is relatively large. We have also proposed several randomization techniques (Supplementary Material, \ref{sec: ablation_data_gen}) in data generation which can further improve model's robustness to error and generalization.

%===============================================================================

\section{Conclusion}
\label{sec:conclusion}

We present CNS that encodes keypoints correspondence into a graph and empoly GNN as neural policy to achieve generalizable visual servoing with high precision and large convergence basin. Even keypoints are intermittent or partially mismatched, CNS is less dependent on the quality of correspondence and achieves higher convergence and precision than conventional methods \cite{chaumette2006visual}. % \cite{chaumette2006visual,corke2001new,allibert2010predictive}. 
When compared with recently popular learning based methods \cite{bateux2018training,felton2021siame}, CNS has better generalization performance because the explicit guidance suppresses the correlation between image appearance and the policy. The training of CNS is independent of specific scenes and can be completely finished in simulation with randomized 3D points. CNS can be directly transferred to real-world scenes without any fine-tuning. The servo of real scenes achieves \textless 0.3° and sub-millimeter servo precision and runs in real-time, which would be a feasible solution to general high precision image servoing tasks.

%===============================================================================

% \clearpage
% The acknowledgments are automatically included only in the final and preprint versions of the paper.
%\acknowledgments{If a paper is accepted, the final camera-ready version will (and probably should) include acknowledgments. All acknowledgments go at the end of the paper, including thanks to reviewers who gave useful comments, to colleagues who contributed to the ideas, and to funding agencies and corporate sponsors that provided financial support.}

%===============================================================================

\balance
\bibliographystyle{IEEEtran}
\bibliography{IEEEabrv,icra2024}

\clearpage
% The acknowledgments are automatically included only in the final and preprint versions of the paper.
\section{Supplementary Material}

\subsection{Hyper-parameters Used in Data Generation}
\label{sec: values_of_hyper-params}
% Here we list the exact values or distributions of hyper-parameters used to sample points and poses, with underlined text as reasons for chosen these values.
Here we list the exact values or distributions of hyper-parameters used to sample points and poses. % These parameters are chosen to be listed values to fully explore the sampling space but also ensure most points can be observed from sampled initial poses and desired poses.

\textbf{Keypoints Distribution:} Empirically, the distribution of keypoints inversely projected to objects in 3D space is a union of a multiple bounded uniform distribution. Denoting $\mathcal{S}(n, x, y, z, h, a, b)$ as a sampling scheme to uniformly sample $n$ points in an elliptic cylinder located at center $(x, y, z)$ with height $h$, major axis $a$ and minor axis $b$. Given total number of points 
% \ul{$N \sim \mathcal{U}(4, 512)$ (although in some texture-rich scenes, some detectors may give over 1000 keypoints and matches, model trained with maximum 512 points is capable to deal with dense keypoints scenes)}
$N \sim \mathcal{U}(4, 512)$ (although in some texture-rich scenes, some detectors may give over 1000 keypoints and matches, model trained with maximum 512 points is capable to deal with dense keypoints scenes)
in scene with maximum size 
% \ul{$r=0.2{\rm m}$ (this value could ensure that most generated points can be observed from a camera which is around $0.5\sim0.9{\rm m}$ away from the scene center with camera intrinsic of $f_x = f_y = 540{\rm mm}, c_x = 320{\rm mm}, c_y = 240{\rm mm}$, shown in Fig.{\ref{fig: data_gen}} and the left panel of Fig.{\ref{fig: exp_setup}})}
$r=0.2{\rm m}$ (this value could ensure that most generated points can be observed from a camera which is around $0.5\sim0.9{\rm m}$ away from the scene center with camera intrinsic of $f_x = f_y = 540{\rm mm}, c_x = 320{\rm mm}, c_y = 240{\rm mm}$, shown in Fig.{\ref{fig: data_gen}} and the left panel of Fig.{\ref{fig: exp_setup}})
, we first determine the number of clusters to generate as 
% \ul{ $N_c \sim \mathcal{U}(3, 1+\log_{2}{N})$ (note $1+\log_{2}{4}=3$ and $1+\log_{2}{512}=10$, we use logarithm to avoid generating too much clusters)}.
$N_c \sim \mathcal{U}(3, 1+\log_{2}{N})$ (note $1+\log_{2}{4}=3$ and $1+\log_{2}{512}=10$, we use logarithm to avoid generating too much clusters).
Each cluster $i$ is randomly assigned with $n_i$ points. Note that $n_i \geq 1$ and $\sum_i n_i \leq 0.8N$, because we preserve at least 20\% points for uniform distribution in the whole scene. Clusters' center coordinates $\{(x_i^c, y_i^c, z_i^c)\}_{i=1}^{N_c}$ are sampled from 
% \ul{$\mathcal{S}(N_c, 0, 0, 0, 0.5r, r, r)$ (here we choose $h=0.5$, allowing points have different depths, i.e., $z$-values)}.
$\mathcal{S}(N_c, 0, 0, 0, 0.5r, r, r)$ (here we choose $h=0.5$, allowing points have different depths, i.e., $z$-values).
Afterwards, points in $i$-th cluster are generated: 
% \ul{ $\mathbf{P}_i = \{(x_j, y_j, z_j)\}_{j=1}^{n_i} = \mathcal{S}(n_i, x_i^c, y_i^c, z_i^c, 0.1r, a, b)$, where $a \sim \mathcal{U}(\frac{0.3r}{\log_{2}{N_c}}, \frac{1.2r}{\log_{2}{N_c}})$ and $b = \frac{1.5r}{\log_{2}{N_c}} - a$ ($h=0.1r$ results in a thin elliptic cylinder, which is more like an elliptic plane mimicking an object surface. Parameters $a, b$ are empirical to generate neither too sparse or too crowded points in single cluster)}.
$\mathbf{P}_i = \{(x_j, y_j, z_j)\}_{j=1}^{n_i} = \mathcal{S}(n_i, x_i^c, y_i^c, z_i^c, 0.1r, a, b)$, where $a \sim \mathcal{U}(\frac{0.3r}{\log_{2}{N_c}}, \frac{1.2r}{\log_{2}{N_c}})$ and $b = \frac{1.5r}{\log_{2}{N_c}} - a$ ($h=0.1r$ results in a thin elliptic cylinder, which is more like an elliptic plane mimicking an object surface. Parameters $a, b$ are empirical to generate neither too sparse or too crowded points in single cluster).
Random rotation around cluster's center is applied to all points belonging to the that cluster to mimic the various object surfaces (with different directions of normals) in the scene. Finally, the residual points are uniformly distributed in the whole scene: $\mathbf{P}_0 = \mathcal{S}(N -\sum_i n_i, 0, 0, 0, 0.5r, r, r)$. The overall 3D points set is the union: $\mathbf{P} \in \mathbb{R}^{N \times 3} = \mathbf{P}_0 \cup \left(\bigcup_{i=1}^{N_c} \mathbf{P}_i\right)$. Given camera's extrinsic $^c_w \mathbf{T}$ and intrinsic $\mathbf{K}$, keypoints in normalized camera plane $\mathbf{S} \in \mathbb{R}^{N \times 2}$ can be obtained by the projection process $\mathbf{S} = {\rm Project}(\mathbf{K}, \ ^c_w \mathbf{T}, \ \mathbf{P})$.

\textbf{Poses Distribution in Simulation:} As shown in Fig.\ref{fig: data_gen}, the camera pose is represented as red-green-blue axes, with $d$ as the distance from camera position to scene center and $\theta$ as angle between vector $\overrightarrow{OP}$ and $xOy$ plane of world coordinate. $d$ ranges from 0.5m to 0.9m; $\theta$ ranges from 30$^\circ$ to 90$^\circ$ for initial poses, and ranges from 70$^\circ$ to 90$^\circ$ for desired poses. After determining the position of camera pose, we sample the rotation of camera pose. Firstly, we make the $z$-axis of camera pose pointing towards scene center $O$, and the $x$-axis parallel to world's $xOy$ plane. We then apply the right hand transform $^{c_t}_{c'_t}\mathbf{R}$ to perturb the sampled rotation with $\mathbf{a}_{\rm max} = [10^\circ, 10^\circ, 60^\circ]$ for initial poses and $\mathbf{a}_{\rm max} = [5^\circ, 5^\circ, 15^\circ]$ for desired poses ($\mathbf{a}$ is the axis-angle representation of $^{c_t}_{c'_t}\mathbf{R}$).

\textbf{Poses Distribution in Real-world:} We use the same ranges of $\theta$ and $\mathbf{a}$ as those in simulation to sample poses and apply perturbance, while $d$ ranges differently due the limited working space of robot arm. In real-world experiments, $d$ ranges from 0.4m to 0.5m for initial poses, and ranges from 0.25m to 0.3m for desired poses.

\subsection{Structure of GConvGRU}
\label{subsec: strcuture_of_gconvgru}
We propose graph convolutional GRU (GConvGRU, denote as $\mathbf{h}_i' = {\rm GConvGRU}(\mathbf{h}_j, \mathbf{h}_i, \mathbf{x}_j, \mathbf{x}_i, \mathbf{p}_j, \mathbf{p}_i)$) built upon PERConv and GRU as:

\begin{footnotesize}
\vspace{-0.3cm}
\begin{equation}
\label{GConvGRU}
    % \left\{
    \begin{aligned}
        \mathbf{z}_i &= {\rm sigmoid}({\rm PERConv}(\textbf{x}_j \ || \ \mathbf{h}_j, \textbf{x}_i \ || \ \mathbf{h}_i, \mathbf{p}_j, \mathbf{p}_i)) \\
        \mathbf{r}_i &= {\rm sigmoid}({\rm PERConv}(\textbf{x}_j \ || \ \mathbf{h}_j, \textbf{x}_i \ || \ \mathbf{h}_i, \mathbf{p}_j, \mathbf{p}_i)) \\
        % \tilde{\mathbf{h}}_i &= {\rm tanh}({\rm PERConv}(\textbf{x}_j \ || \ (\mathbf{r}_j \odot \mathbf{h}_j), \textbf{x}_i \ || \ (\mathbf{r}_i \odot \mathbf{h}_i), \\
        %     & \qquad \qquad \qquad \qquad \ \ \mathbf{p}_j, \mathbf{p}_i)) \\
        \tilde{\mathbf{h}}_i &= {\rm tanh}({\rm PERConv}(\textbf{x}_j \ || \ (\mathbf{r}_j \odot \mathbf{h}_j), \textbf{x}_i \ || \ (\mathbf{r}_i \odot \mathbf{h}_i), \mathbf{p}_j, \mathbf{p}_i)) \\
        \mathbf{h}_i' &= (1 - \mathbf{z}_i) \odot \mathbf{h}_i + \mathbf{z}_i \odot \tilde{\mathbf{h}}_i
    \end{aligned}
    % \right.
\end{equation}
\end{footnotesize}

\noindent where $\odot$ is the element-wise production. Our GConvGRU is different from gated graph convolution proposed by \cite{li2015gated}. The latter runs graph convolution and GRU cell sequentially, while in our GConvGRU, graph convolution directly participates in the predictions of update gate $\mathbf{z}_i$, reset gate $\mathbf{r}_i$ and candidate hidden state $\tilde{\mathbf{h}}_i$.

\subsection{Network Training}
\begin{figure}[htbp]
\includegraphics[width=0.48\textwidth]{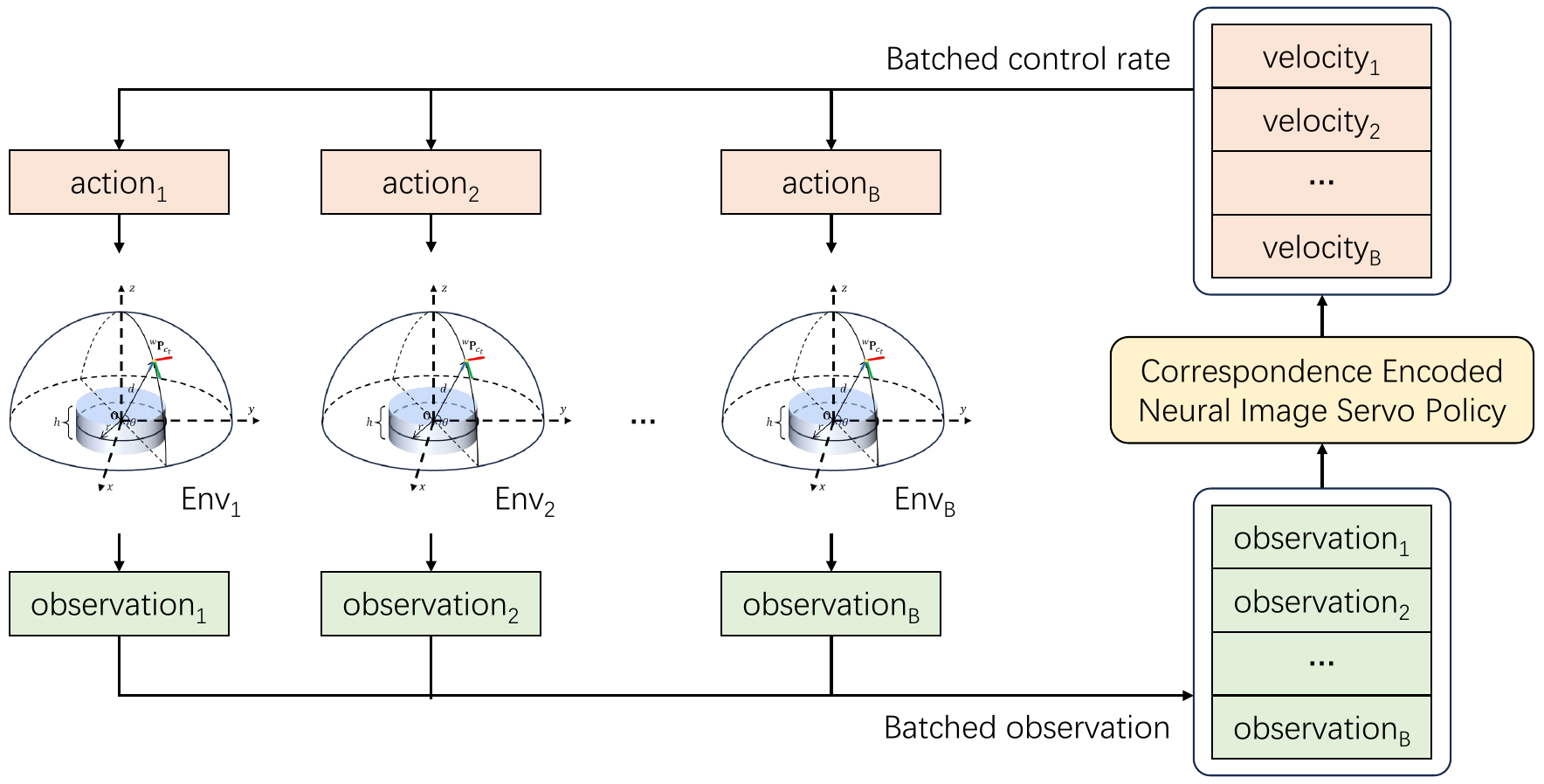}
\centering
\caption{CNS takes the batched observation (B=64 in this work) and predicts batched velocity control rates. Cameras in environments take the corresponding velocity control rates and move to the next poses. Data is always collected with the latest servo policy.}
\label{fig: training}
\end{figure}

When training CNS, we launch 64 simulation environments simultaneously sampling points and projecting them to the normalized image plane as observations (Fig.\ref{fig: training}). Predictions of CNS are fed back to cameras in environments to update their poses. This means data is always collected with the latest servo policy. We resample the points, initial and desired poses for an environment if the camera in that environment reaches the desired pose or is ill-posed (too close or too far away, or most keypoints are out of camera's FoV).

\subsection{Experimental Setup}
\label{subsec: exp_setup}

\begin{figure*}[t]
\includegraphics[width=\textwidth]{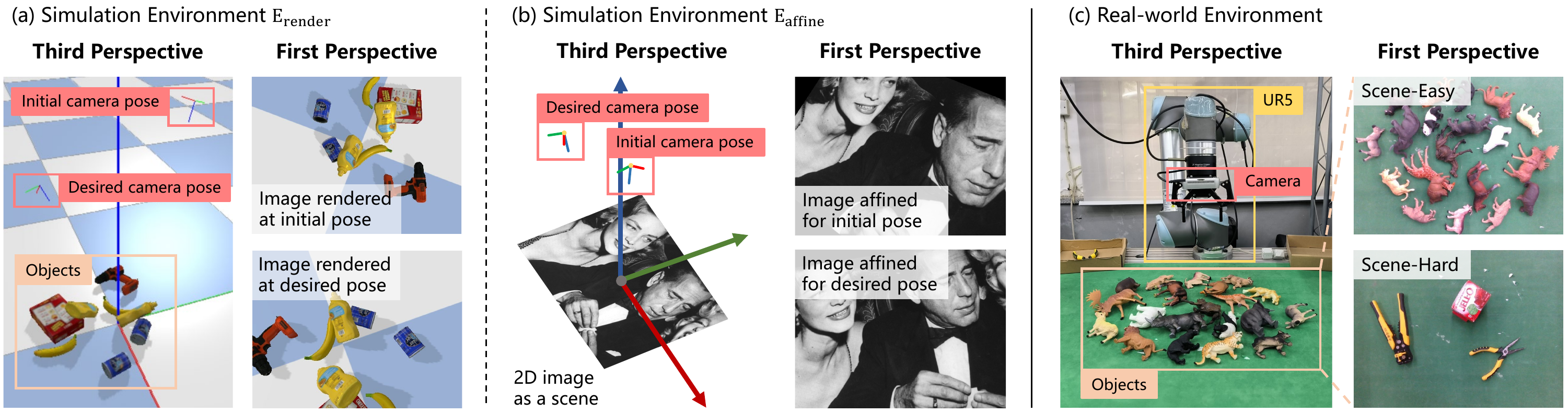}
\centering
\caption{Simulation (a\&b) and real-world (c) experiments setup. (a) Simulation environment ${\rm E}_{\rm render}$: total 150 scenes (8$\sim$12 objects per scene) and 150 pairs of initial-desired poses (one pair per scene). Camera intrinsic: $f_x$=$f_y$=540mm, $c_x$=320mm, $c_y$=240mm. (b) Simulation environment ${\rm E}_{\rm affine}$: one fixed 2D image as a scene, total 150 pairs of initial-desired poses. Same camera intrinsic as that of ${\rm E}_{\rm render}$. (c) Real-world environment: total 2 scenes and 60 pairs of initial-desired poses, Scene-Easy (20 objects) shares the same sampled poses with Scene-Hard (3 objects). Calibrated camera intrinsic: $f_x$=$f_y$=615.7mm, $c_x$=315.8mm, $c_y$=248.1mm.
}
\label{fig: exp_setup}
\end{figure*}

As shown in Fig.\ref{fig: exp_setup}a, we generate total 150 scenes in simulation environment ${\rm E}_{\rm render}$ and randomly sample an initial pose and a desired pose for each scene. Each scene is initialized with 8 to 12 randomly placed objects. Image observed from specific pose is rendered by PyBullet simulation engine. When compared with \cite{bateux2018training,felton2021siame}, we use simulation environment ${\rm E}_{\rm render}$ (Fig.\ref{fig: exp_setup}b). We use the same image as they do. We sampled total 150 pairs of initial-desired poses for evaluation.

We benchmark methods in two real-world scenes: Scene-Easy and Scene-Hard (Fig.\ref{fig: exp_setup}c). We uniformly scatter 20 objects (this results in rich keypoints) in Scene-Easy, and randomly sample 60 initial-desired pose pairs for statistics. In Scene-Hard, there are 3 objects, and we use the same initial-desired pose pairs as those in Scene-Easy, but objects are often partially observed in the initial or desired pose. Moreover, the red box in Scene-Hard is placed higher than the other two yellow pliers.

% \begin{figure}[htbp]
% \includegraphics[width=\linewidth]{figures/exp_setup.pdf}
% \centering
% \caption{Simulation (left) and real-world (right) experiments setup. Simulation: total 150 scenes (8$\sim$12 objects per scene) and 150 pairs of initial-desired poses (one pair per scene); Camera intrinsic: $f_x$=$f_y$=540mm, $c_x$=320mm, $c_y$=240mm. Real world: total 2 scenes and 60 pairs of initial-desired poses, Scene-Easy (20 objects) shares the same sampled poses with Scene-Hard (3 objects). Calibrated camera intrinsic: $f_x$=$f_y$=615.7mm, $c_x$=315.8mm, $c_y$=248.1mm.
% }
% \label{fig: exp_setup}
% \end{figure}

\subsection{Ablation Study on Observer}
Different keypoint detector and feature matching methods give different number, distribution and position precision of keypoints. Results from observer will influence the graph structure, servo precision and servo rate. We compare several observers, including ORB, SIFT, AKAZE, BRISK and SuperGlue \cite{sarlin2020superglue} (a learning-based feature matching method built upon SuperPoint \cite{detone2018superpoint}). As shown in Table \ref{tab: ablation_frontend}, AKAZE and SuperGlue give the best success ratio and SuperGlue also gives farthest convergence steps. ORB is the fastest of each frame, SIFT results in highest servo precision. 

% BRISK is normal at each metric.

% \begin{table}[htbp]
\begin{table*}[b]
\centering
\caption{Performance with different observers.}
\label{tab: ablation_frontend}
% \resizebox{\linewidth}{!}{
% \resizebox{\textwidth}{!}{
\begin{tabular}{l|l|l|l|l|l|l|l}
\hline
          & SR(\%) (95\% ci)   & TS (0.04s)  & RE (°)      & TE (mm)     & mOT (ms)   & mCT (ms)   & mTT (ms) \\ \hline
AKAZE     & \textbf{100}       & 207.8± 43.8 & 0.115±0.111 & 1.394±1.402 & 45.09±3.48 & 11.51±1.65 & 56.60±3.87    \\ \hline
ORB       & 98.67 (96.83, 100) & 194.4± 55.3 & 0.226±0.199 & 2.873±2.821 & \textbf{14.20}±0.62 & 11.33±1.33 & \textbf{25.62}±1.48    \\ \hline
BRISK     & 98.67 (96.83, 100) & 206.3± 52.3 & 0.134±0.127 & 1.677±1.625 & 37.47±10.2 & 12.34±1.59 & 49.81±10.6    \\ \hline
SIFT      & 99.33 (98.03, 100) & 203.2± 43.1 & \textbf{0.109}±0.102 & \textbf{1.379}±1.263 & 68.89±7.02 & 12.65±1.59 & 81.53±7.76    \\ \hline
SuperGlue & \textbf{100}       & \textbf{183.9}± 81.8 & 0.417±0.297 & 5.707±4.436 & 79.19±3.51 & \textbf{9.03}±0.37 & 88.22±3.73    \\ \hline
\end{tabular}
% }
\end{table*}

\subsection{Ablation Study on Data Generation}
\label{sec: ablation_data_gen}
As discussed in section \ref{sec: data_generation}, several techniques are proposed to produce realistic keypoint distribution and simulate the time-varying behavior of keypoints. We would briefly reintroduce them as:
\begin{itemize}
    \item KM (Keypoint Mismatch): Augmentation that randomly mismatch 10\% keypoints;
    \item UKD (Uniform Keypoint Dropout): Augmentation that randomly drop 10\% keypoints (with equal probability) in the graph;
    \item WKD (Weighted Keypoint Dropout): Similar as UDK, but the drop probability is weighted according to observable region.
    \item UK (Uniform Keypoint Distribution): Keypoint sampling method that uniformly scatter points in 3D space.
    \item CK (Clustered Keypoint Distribution): Keypoint sampling method that scatter points in clusters.
\end{itemize}

\begin{table}[!h]
\centering
\caption{Effects on data generation techniques.}
\label{tab: ablation_datagen}
\resizebox{\linewidth}{!}{
\begin{tabular}{l|l|l|l|l}
\hline
          & SR(\%) (95\% ci)     & TS (0.04s)  & RE (°)      & TE (mm) \\ \hline
CK        & 95.33 (91.96, 98.71) & 269.6±133.9 & 0.159±0.228 & 1.769±2.224 \\ \hline
UK+UKD+KM & 96.67 (93.79, 99.54) & \textbf{192.0}± 53.0 & 0.132±0.137 & 1.619±1.752 \\ \hline
UK+WKD+KM & 98.00 (95.76, 100)   & 218.4±110.3 & 0.130±0.132 & 1.512±1.630 \\ \hline
CK+UKD+KM & 99.33 (98.03, 100)   & 203.8± 62.9 & 0.125±0.105 & 1.528±1.323 \\ \hline
CK+WKD+KM & \textbf{100}         & 207.8± 43.8 & \textbf{0.115}±0.111 & \textbf{1.394}±1.402 \\ \hline
\end{tabular}
}
\end{table}

As shown in Table \ref{tab: ablation_datagen}, since the raw keypoint correspondences are intermittent and sometimes incorrect, to achieve direct sim-to-real transfer, we should ensure CNS to have the ability to handle intermittent and incorrect correspondence. From the first line of the table, CNS trained with perfect keypoint correspondences can only achieve a convergence rate of 95.33$\%$. After adding KM, the performance of the model is improved, which shows that KM is important for generalization and mimics the correspondence quality in actual use to enable trained policy to handle mismatches. The combination of WKD, CK and KM achieves the highest convergence rate and precision, which is more close to the distribution and time varing behavior of keypoints in real-world scenes.

\subsection{Ablation Study on Velocity Design}
\label{sec: ablation_on_dd_vel}

\begin{figure}[tb]
\includegraphics[width=0.48\textwidth]{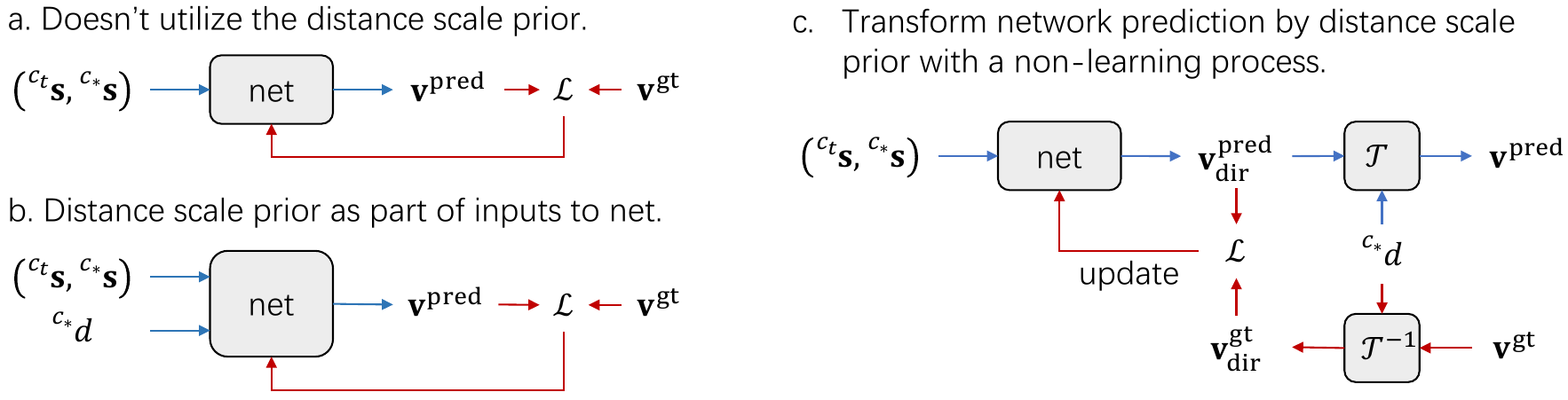}
\centering
\caption{Different network objectives design.}
\label{fig: ablation_distance_decouple}
\end{figure}

We evaluate the effect of distance decoupled velocity design. Two extra network are trained, with the first one (Fig.\ref{fig: ablation_distance_decouple}a) doesn't utilize the distance prior at all and the second one (Fig.\ref{fig: ablation_distance_decouple}b) takes the distance prior as part of input features (CNS in main text adopts distance decoupled design as shown in Fig.\ref{fig: cns} and Fig.\ref{fig: ablation_distance_decouple}c). The distance from camera center to scene center in original environment ${\rm E}_{\rm render}$ ranges from 0.5m to 0.9m with mean=0.711m and std=0.118m, we reduce or enlarge the size of objects in scene and distance from camera to scene center simultaneously for 5 times (denote as x0.2 and x5.0 in Table \ref{tab: ablation_distance_decouple}, respectively). In scale x0.2 and x5.0, model using distance decoupled velocity achieves comparable performance in SR, TS and RE with those in x1.0 (the same distance scale in training), while TE is reduced or enlarged by about 5 times respectively, as expected. Model not knowing distance prior predict the same control rate as in scale x1.0, since the rendered image are almost the same if both the distance and object size are scaled by same times. Therefore, in scale x0.2, the controller moves relatively too fast that objects are easy to be out of camera's view, resulting lower success ratio. The model also jitters around the desired pose and uses more time steps to convergence. While in scale x5.0, the model moves relatively too slow to reach the surroundings of desired pose, which also increases convergence time steps. As for model taking distance prior as part of input features, however, performs even worse in scale x0.2 compared with model not knowing distance prior. This indicates model trained with distance prior as features can not generalize to other distance scales that not saw in training set.

\begin{table}[tb]
\centering
\caption{Effect of distance decoupled velocity design.}
\label{tab: ablation_distance_decouple}
\resizebox{\linewidth}{!}{
\begin{tabular}{l|l|l|l|l|l}
\hline
                                         & scale & SR(\%) (95\% ci)     & TS (0.04s)  & RE (°)      & TE (mm)     \\ \hline
\multirow{3}{*}{w/o distance prior}      & x0.2  & 74.00 (66.98, 81.02) & 446.9±177.3 & 0.198±0.360 & 0.502±0.813 \\ \cline{2-6} 
                                         & x1.0  & 99.33 (98.03, 100)   & 228.8±113.5 & 0.168±0.274 & 2.053±4.406 \\ \cline{2-6} 
                                         & x5.0  & 98.00 (95.76, 100)   & 453.9±115.7 & 0.153±0.124 & 12.92±14.44 \\ \hline
\multirow{3}{*}{distance prior as input} & x0.2  & 61.33 (53.54, 69.13) & 376.0±152.6 & 0.172±0.222 & 0.416±0.604 \\ \cline{2-6} 
                                         & x1.0  & \textbf{100}         & \textbf{203.7}± 53.3 & 0.142±0.137 & 1.820±1.903 \\ \cline{2-6} 
                                         & x5.0  & \textbf{100}         & 412.4±143.7 & 0.289±0.148 & 19.91±12.34 \\ \hline
\multirow{3}{*}{distance decoupled}      & x0.2  & \textbf{98.67} (96.83, 100) & \textbf{205.6}± 40.5 & \textbf{0.123}±0.135 & \textbf{0.308}±0.370 \\ \cline{2-6} 
                                         & x1.0  & \textbf{100}         & 207.8± 43.8 & \textbf{0.115}±0.111 & \textbf{1.394}±1.402 \\ \cline{2-6} 
                                         & x5.0  & \textbf{100}         & \textbf{203.0}± 39.5 & \textbf{0.122}±0.133 & \textbf{7.544}±9.585 \\ \hline
\end{tabular}
}
\end{table}

\subsection{Impact of Imprecise Distance Prior Estimation}
\label{subsec: distance_estimation}
Previous experiments use the ground truth distance prior. However, in real world applications, users may not have a depth sensor to obtain this, they may just give a rough estimation of distance. As mentioned above, the mean distance between camera and scene center of total 150 scenes is 0.711m. We suppose the distance prior is always estimated to be 0.25m, 0.5m, 1m and 2m, and evaluate performance again to see the impacts of imprecise distance estimation. Results (Table \ref{tab: distance_estimation}) show that when distance is underestimated, the controller takes longer time steps to reach the desired pose, but the success ratio and servo precision are not affected. While overestimate distance will success less as objects may easily move out of view. The controller will also jitters around the desired pose and takes more time steps to convergence. However, it doesn't loose much servo precision in the success cases.

\begin{table}[htbp]
\centering
\caption{Impact of imprecise distance prior estimation.}
\label{tab: distance_estimation}
\resizebox{\linewidth}{!}{
\begin{tabular}{l|l|l|l|l}
\hline
                                    & SR(\%) (95\% ci)     & TS (0.04s)  & RE (°)      & TE (mm) \\ \hline
% use gt: ${\rm mean}(d^{\rm gt})$= 0.711m     & 100                  & 207.8± 43.8 & 0.115±0.111 & 1.394±1.402 \\ \hline
use gt: $\hat{d}=d^{\rm gt}$     & 100                  & 207.8± 43.8 & 0.115±0.111 & 1.394±1.402 \\ \hline
estimation: $\hat{d} \equiv$ 0.25m  & 100                  & 432.9±110.0 & 0.100±0.077 & 1.240±1.201 \\ \hline
estimation: $\hat{d} \equiv$ 0.50m  & 100                  & 251.4± 62.1 & 0.121±0.136 & 1.454±1.611 \\ \hline
estimation: $\hat{d} \equiv$ 1.00m  & 98.67 (96.83, 100)   & 217.5± 56.1 & 0.125±0.161 & 1.556±2.032 \\ \hline
estimation: $\hat{d} \equiv$ 2.00m  & 80.00 (73.60, 86.40) & 291.2±105.2 & 0.105±0.085 & 1.361±1.170 \\ \hline
\end{tabular}
}
\end{table}

\end{document}